\documentclass{article}

\usepackage{PRIMEarxiv}

\usepackage[utf8]{inputenc} % allow utf-8 input
\usepackage[T1]{fontenc}    % use 8-bit T1 fonts
\usepackage{hyperref}       % hyperlinks
\usepackage{url}            % simple URL typesetting
\usepackage{booktabs}       % professional-quality tables
 \usepackage{ifthen}
\usepackage{amsfonts}       % blackboard math symbols
\usepackage{nicefrac}       % compact symbols for 1/2, etc.
\usepackage{microtype}      % microtypography
\usepackage{lipsum}
\usepackage{fancyhdr}       % header
\usepackage{graphicx}       % graphics
\graphicspath{{media/}}     % organize your images and other figures under media/ folder

\usepackage{graphicx} % DO NOT CHANGE THIS
\usepackage{hyperref}
\usepackage{algorithm}
\usepackage{algorithmic}
\usepackage{subcaption}
\usepackage{multirow}
\usepackage{multicol}
\usepackage{amsmath} %added 
\usepackage{array} %added 
\usepackage{verbatim}%added 
\usepackage{appendix}%added 

\usepackage{listings}
\usepackage{xcolor}

\definecolor{codegreen}{rgb}{0,0.6,0}
\definecolor{codegray}{rgb}{0.5,0.5,0.5}
\definecolor{codepurple}{rgb}{0.58,0,0.82}
\definecolor{backcolour}{rgb}{0.95,0.95,0.92}

\lstdefinestyle{mystyle}{
    backgroundcolor=\color{backcolour},   
    commentstyle=\color{codegreen},
    keywordstyle=\color{magenta},
    numberstyle=\tiny\color{codegray},
    stringstyle=\color{codepurple},
    basicstyle=\ttfamily\footnotesize,
    breakatwhitespace=false,         
    breaklines=true,                 
    captionpos=b,                    
    keepspaces=true,                 
    numbers=none,                    
    numbersep=5pt,                  
    showspaces=false,                
    showstringspaces=false,
    showtabs=false,                  
    tabsize=2
}
\title{Physical Rule-Guided Convolutional Neural Network
}

\author{
 Kishor Datta Gupta \\
 Cyber Physical Systems \\
 Clark Atlanta University \\
 Atlanta, GA\\
 \texttt{kgupta@cau.edu} \\
 \And
 Marufa Kamal \\
 Dept. of CSE \\
 BRAC University \\
 Dhaka, Bangladesh\\
 \texttt{marufa.kamal1@g.bracu.ac.bd} \\
 \And
 Rakib Hossain Rifat \\
 Dept. of Computer Science \\
 Texas Tech University \\
 Lubbock, TX 79409\\
 \texttt{rrifat@ttu.edu} \\
 \And
 Mohd Ariful Haque \\
 Cyber Physical Systems \\
 Clark Atlanta University \\
 Atlanta, GA\\
 \texttt{mohdariful.haque@students.cau.edu} \\
 \And
 Roy George \\
 Cyber Physical Systems \\
 Clark Atlanta University \\
 Atlanta, GA\\
 \texttt{rgeorge@cau.edu} \\
}

\begin{document}
\maketitle

\begin{abstract}
The black-box nature of Convolutional Neural Networks (CNNs) and their reliance on large datasets limit their use in complex domains with limited labeled data. Physics-Guided Neural Networks (PGNNs) have emerged to address these limitations by integrating scientific principles and real-world knowledge, enhancing model interpretability and efficiency. This paper proposes a novel Physics-Guided CNN (PGCNN) architecture that incorporates dynamic, trainable, and automated LLM-generated, widely recognized rules integrated into the model as custom layers to address challenges like limited data and low confidence scores. The PGCNN is evaluated on multiple datasets, demonstrating superior performance compared to a baseline CNN model. Key improvements include a significant reduction in false positives and enhanced confidence scores for true detection. The results highlight the potential of PGCNNs to improve CNN performance for broader application areas.
\end{abstract}

% keywords can be removed

\section{Introduction}

\begin{figure}[htbp]
    \centering
    \includegraphics[ height = 4.5cm]{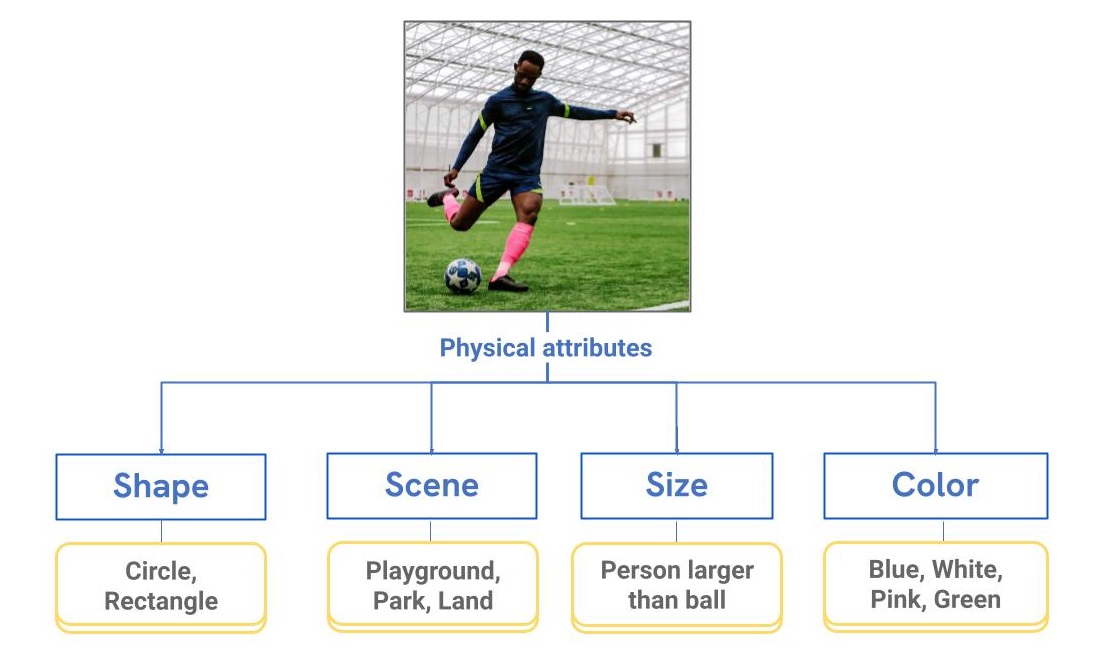}
    \caption{ Physical Attributes of a Ball and Player in an Image Context}
    \label{fig:Introduction Example}
\end{figure}

Convolutional Neural Networks (CNNs) have become a transformative tool across a wide range of fields, celebrated for their ability to automatically learn and extract features from raw data. This architecture supports a diverse range of applications, including computer vision tasks such as image classification, object detection, and segmentation, as well as broader domains like natural language processing and genomics, where it has substantially advanced the state of the art. Despite its remarkable success, the black-box nature of CNN models and their reliance on large labeled datasets pose significant challenges, particularly in complex domains with limited data availability\cite{fu2023high}, \cite{isufi2019generalizing}.
To address these challenges, researchers have begun to explore the integration of domain-specific knowledge and scientific principles into neural network architectures, leading to the development of Physics-Guided Neural Networks (PGNNs). PGNNs represent a shift from purely data-driven approaches to models that incorporate physical laws, scientific conditions, real-world constraints, and common-sense reasoning into the learning process\cite{yaman2020self}, \cite{yaman2021ground} \cite{yaman2022self}. Incorporating physical attributes into the CNN architecture can enhance the model's performance and interpretability. Additionally, incorporating physical loss into the loss functions can lead to more effective model training\cite{huang2022physics}. Figure \ref{fig:Introduction Example}, highlights some physical attributes a CNN might use to detect both a ball and a player from an image. Attributes like shape, the surrounding scene, size comparisons, and color can help the model distinguish these objects, enhancing the accuracy and reliability of detection by interpreting attributes in a manner similar to the human mind. Although the figure portrays an image input, similar principles can be applied to text\cite{clifton2001sentence}\cite{biber2019text} or sound inputs\cite{apple1979effects}, where attributes such as tone, context, semantics, or linguistic structure play a crucial role in decision-making.

This paper introduces a novel Physics-Guided Convolutional Neural Network(PGCNN) architecture that leverages the strengths of CNNs while incorporating dynamic, trainable, and automated custom layers based on physical rules to address challenges such as limited data availability and low confidence scores, thereby increasing the model's trustworthiness. The PGCNN framework is presented using object detection as an illustrative case study to demonstrate its potential and additionally harness knowledge embedded within Large Language Models(LLMs). We aim to improve performance by reducing false positives, enhancing model performance, managing object size variation, and considering physical context. The PGCNN is evaluated on multiple datasets and custom rules are curated incorporating them into the Faster R-CNN model using a CNN architecture, ResNet-50. The datasets consist of vehicle classes in land and water landscapes. The initial custom layers are designed to remove bounding boxes during training that are fully or partially inside another box of the same class, reducing false positives. They also eradicate bounding boxes that are fully contained within or exhibit significant overlap with another box of the same class. The following custom layers consider the physical context surrounding the vehicle by examining whether the majority of the scene is composed of landmass or watermass, providing additional cues for improved detection. The final custom rule leverages common sense and human perceptual principles by comparing the relative sizes of different vehicles and employing OpenAI LLM dynamically to generate weight relationships among the vehicle classes with respect to their size.

The proposed PGCNN framework being evaluated on a multi-environmental dataset allows the model to be trained and tested across diverse scenic contexts. While the PGCNN and the baseline Faster R-CNN ResNet-50 models may exhibit similar mean Average Precision (mAP) scores, the PGCNN demonstrates significant improvements in inference results, showing a notable reduction in false positives and an increase in confidence scores for true detections, highlighting its potential to enhance the trustworthiness and reliability of object detection in diverse application areas. Moreover, PGCNN can be designed to be adaptable across diverse domains. By incorporating custom rules and domain-specific knowledge, researchers can explore and extend this approach to tackle challenges in various applications, making it a versatile tool for enhancing model reliability.

\section{Literature Review}

In recent years, a lot of work has been directed towards the development of Physics-Guided Neural Networks (PGNNs) and covering various related tasks. In this section, we briefly describe some of the related works on this.
In the year 2022, \cite{huang2022physics} developed a physics-guided deep neural network (PGDNN) that uses 5,400 labeled and 1,440 unlabeled data. By combining neural networks with finite element models with achieved over 80\% accuracy in identifying structural damage. Moreover, they suggested exploring its use in machine monitoring, visual recognition, and bridge damage detection. Similarly, in a different study \cite{zhang2019pgnn} presents a Physics-Guided Neural Network (PGNN) for fourier ptychographic microscopy (FP) in biological data. Their model exploratory analysis used real and simulated datasets with peak signal-to-noise ratio (PSNR), and structural similarity index (SSIM). Their model outperforms explanation focus and PGNN improves better than ePIE in high-defocus and high-exposure conditions.
 
Another work by \cite{banerjee2023physics} primarily focuses on reviewing existing methodologies over 250 papers on physics-informed computer vision (PICV), covering models like PI Reinforcement Learning and Physics-Incorporated Safety Prediction. They suggested the most effective model is PICV. In another research \cite{ling2024physics} suggested that PINNs and physics-guided nnU-Net are better models for blood flow in iVFM than the original method by using Doppler and CFD simulations. With similar relation to our work, \cite{li2023physics} have experimented with a hybrid model combining CNN and PGNN to diagnose sensor issues in aero-engines. In another work, researcher \cite{fernandez2023physics} experimented with the PG-BNN combined with Bayesian computation (ABC). Their analysis showed that PG-BNN outperformed traditional methods in forecasting the shear strength of reinforced concrete columns. In some object detection papers we have seen researchers focusing on the shortage of data for experimenting such as ship detection resulting to use synthetic images which can be often unreliable with real ocean shots\cite{zhang2023physics}. Moreover, researchers \cite{singh2018dock} detected the transfer of common-sense knowledge on the DOCK technique in the year 2018 with MS COCO dataset achieving a (mAP) score of 21.4\% on their model. Their work integrates visual and semantic similarity with geographical and attribute knowledge to improve identification performance. Inspired by different current researches we propose our PGCNN framework with hopes of more exploration. 

\section{Methodology}

\begin{figure*}[htbp]
    \centering    
    \includegraphics[width=13cm, height=4.4cm]{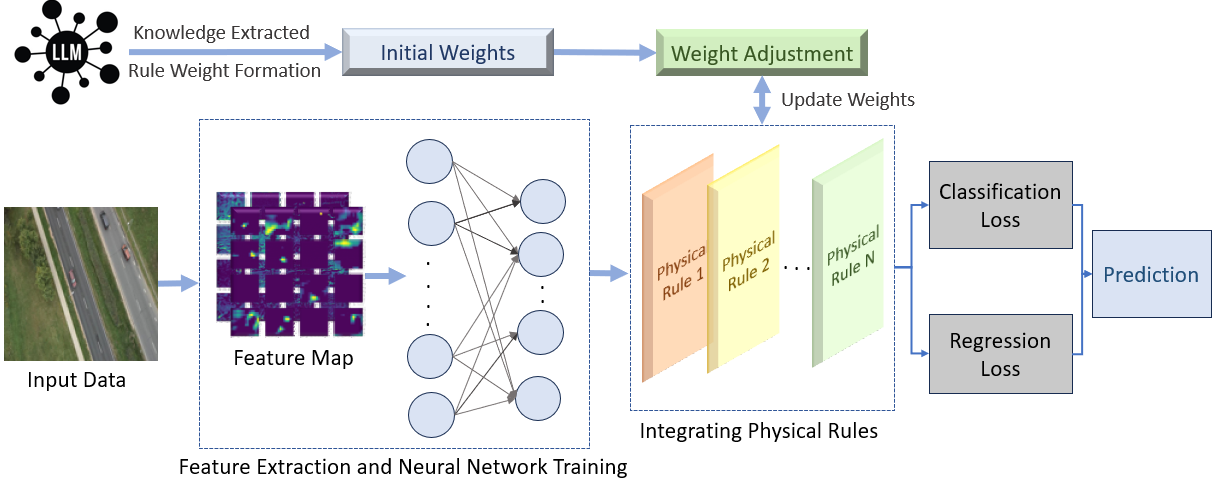}
    \caption{Overview of the PGCNN Framework}
    \label{fig:Architecture}
\end{figure*}

We present an enhanced and modified CNN framework integrating a Neural Network backbone with multiple physical rules integrated into the model as custom layers for improved results. The neural network serves as the baseline model. Figure \ref{fig:Architecture} portrays an overall framework of how custom layers with LLM knowledge can be integrated with a CNN based model.

Our final PGCNN model incorporates a base CNN network and several novel physical rules as custom layers, which are described below:

\subsection{Shape-Based Object Detection Layer }
In real-world scenarios, objects can often be identified by the unique combination of geometric shapes they encompass. For instance, viewing a traditional car from a top-down perspective might reveal the presence of rectangles, squares, trapezoids, and triangles, either individually or in combination. This observation forms the basis for our novel shape detection rule.
It begins by segmenting each detected object into its fundamental positional components. Following this segmentation, we apply shape detection algorithms to identify and count the basic geometric shapes present within each segment.

Let \( S = \{s_1, s_2, \dots, s_n\} \) represent the set of detected shapes, where \( s_i \) denotes the count of a specific shape \( i \) (e.g., rectangle, square, trapezoid, triangle, etc.). To enhance the accuracy of our object detection, we compare the detected shape counts against a pre-established knowledge base \( K \) created by a large language model (LLM). The knowledge base \( K \) contains expected shape counts for various objects, represented as \( K = \{k_1, k_2, \dots, k_n\} \), where \( k_i \) is the expected count of shape \( i \) for a specific object. The confidence \( C \) in the object detection is updated based on the similarity between the detected shape counts \( S \) and the knowledge base \( K \). If the detected values closely match the expected counts in the knowledge base, the confidence score \( C \) increases, indicating a higher likelihood of correct object detection. Conversely, if the detected values deviate significantly from the expected counts, the confidence score decreases, reflecting lower certainty in the detection.

This relationship can be expressed mathematically as:

\begin{equation}
C = 1 - \frac{1}{1 + \exp\left(-\alpha \sum_{i=1}^{n} \left(\frac{s_i - k_i}{k_i}\right)^2 \right)}
\end{equation}

\( \alpha \) is a scaling factor that controls the sensitivity of the confidence adjustment, and the expression inside the exponential function represents the sum of squared relative errors between the detected shape counts and the expected counts in the knowledge base. To ensure that the detected shapes and counts align closely with the knowledge base, the confidence is increased in the correctness of the object detection.

\subsection{Redundancy Elimination Layer}
In our pipeline, a custom layer is employed to refine bounding box proposals by eliminating redundant boxes, reducing false positives, and enhancing detection accuracy. During the forward pass, this layer takes bounding boxes $B \in \mathbb{R}^{n \times 4}$ and their corresponding labels $L \in \mathbb{R}^n$ as input.

The layer iterates over pairs of bounding boxes $(B_i, B_j)$ and compares them using a spatial rule. This rule evaluates whether one box is fully or partially contained within another box of the same class by checking the spatial relationships. We define the bounding boxes as \( B_i = [x_{i1}, y_{i1}, x_{i2}, y_{i2}] \) and \( B_j = [x_{j1}, y_{j1}, x_{j2}, y_{j2}] \). The rule is then:

\begin{equation}
\text{Rule: } \left( x_{i1} \geq x_{j1} \right) \land \left( y_{i1} \geq y_{j1} \right) \land \left( x_{i2} \leq x_{j2} \right) \land \left( y_{i2} \leq y_{j2} \right)
\label{eq:spatial_rule}
\end{equation}

If the condition is met, the index of the redundant box is recorded. After all redundant boxes are identified, their indices are removed.
In our modified pipeline, we incorporate another physical rule to refine bounding box proposals by eliminating redundant boxes, thereby enhancing detection precision. This rule operates by accepting bounding boxes $B \in \mathbb{R}^{n \times 4}$ and corresponding labels $L \in \mathbb{R}^n$ during the forward pass. It iterates over pairs of bounding boxes $(B_i, B_j)$, comparing each to determine if one is fully contained within or significantly overlaps another of the same class. The redundancy check is conducted via the defined physical rule and a redundancy factor (RF) value, which evaluates the spatial relationship by calculating the area $A(B_i)$ of $B_i$ and the intersection area $A(B_i \cap B_j)$. The overlap percentage is given by:

\begin{equation}
\text{Overlap}(B_i, B_j) = \frac{A(B_i \cap B_j)}{A(B_i)}
\label{eq:overlap}
\end{equation}
A box is deemed redundant if:

\begin{equation}
B_i \subseteq B_j \quad \text{or} \quad \text{Overlap}(B_i, B_j) \geq \text{RF}
\label{eq:redundancy_condition}
\end{equation}

Redundant box indices are recorded and removed from $B$ and $L$, with refined boundary boxes.
\subsection{Context-Aware Weight Adjustment Layer (CAWAL)} 
Another addition in the model is the ContextAwareAdjustmentLayer(CAWAL) helps to improve the accuracy of object detection in complex scenes by adjusting logit scores during training based on contextual cues like the presence of relevant environmental features (e.g., cars on roads, boats on water). 
\begin{algorithm}[!htb]
\caption{Scene-Based Contextual Weight Adjustment(CAWAL) Procedure}
\label{alg:cawal}

\textbf{Inputs:}
\begin{itemize}
\item $\mathbf{P} \in \mathbb{R}^{m \times n}$: Tensor of predicted scene labels, where $m$ is the number of scenes and $n$ is the number of possible labels per scene.
\item $\mathbf{L} \in \mathbb{R}^{k}$: Tensor of predicted object labels, where $k$ is the number of objects.
\item $\mathbf{S} \in \mathbb{R}^{k \times c}$: Tensor of logits, where $c$ is the number of classes.
\end{itemize}
\textbf{Parameters:}
\begin{itemize}
\item $\mathbf{A}$: Index corresponding to a specific scene attribute (e.g., Label A).
\item $\mathbf{B}$: Index corresponding to another specific scene attribute (e.g., Label B).
\item $\mathbf{C} \subseteq \mathbb{R}^{v}$: Set of indices corresponding to a specific subset of object labels, where $v$ is the number of these classes.
\item $\alpha \in \mathbb{R}$: Factor for adjusting logits, expressed as a percentage.
\end{itemize}
\textbf{Output:}
\begin{itemize}
\item Adjusted $\mathbf{L}$ and $\mathbf{S}$.
\end{itemize}
\begin{algorithmic}[1]
\STATE Convert $\mathbf{P}$ into a list representation $\mathbf{P}{\text{list}}$.
\STATE Initialize counter $C{A+B} = 0$.
\FOR{each list $\mathbf{p}{i}$ in $\mathbf{P}{\text{list}}$}
\STATE $C_{A+B} = C_{A+B} + \text{count}(\mathbf{p}{i}, \mathbf{A}) + \text{count}(\mathbf{p}{i}, \mathbf{B})$
\ENDFOR
\STATE Compute $N_{\text{total}} = \sum_{i=1}^{m} |\mathbf{p}{i}|$, where $|\mathbf{p}{i}|$ is the length of each list in $\mathbf{P}{\text{list}}$.
\IF{$C{A+B} > \Delta t \times N_{\text{total}}$}
\STATE Set $\beta = \frac{\alpha}{100}$
\FOR{each $i \in {1, 2, \dots, k}$}
\IF{$L_{i} \in \mathbf{C}$}
\STATE $S_{i,j} = S_{i,j} \times (1 + \beta) \quad \forall j \in {1, 2, \dots, c}$
\ENDIF
\ENDFOR
\ENDIF
\RETURN Adjusted $\mathbf{L}$ and $\mathbf{S}$.
\end{algorithmic}
\end{algorithm}

The CAWAL layer is initialized with parameters that include the names of contextual entities, such as various environmental features, and a list of objects corresponding to specific categories. The layer takes a weight adjustment threshold, which determines the extent to which the confidence scores of relevant predictions are adjusted when certain conditions are met.

During the forward pass, the CAWAL layer first converts the predicted scene labels into a list format. It then counts the occurrences of specific contextual labels across all predicted values and calculates the total number of individual labels. For example, the layer may count occurrences of certain labels(A) representing one set of contextual entities (e.g., water mass) and another set(B) representing different entities (e.g., land areas). If the proportion of labels from the first set exceeds an iteratively selected threshold 
\(\Delta t\) (e.g., 30\% of the total labels), the logit scores of associated object categories are increased by a factor($\alpha$) derived from the weight adjustment percentage. Similarly, if the proportion of labels from the second set exceeds the threshold(\(\Delta t\)), the logit scores(S) of another set of object categories with respect to that category are adjusted.

We provide an algorithmic overview \ref{alg:cawal} of our CAWAL Layer, which adjusts logit scores during training based on contextual factors related to the object's surroundings.
\subsection{Hybrid Weight Adjustment Layer (HWAD)}
In our framework, we introduce the final novel weight adjustment layer named HybridWeightAdjustmentLayer (HWAD), designed to adjust detection logits based on predefined size comparison rules stored in a JSON file. This layer enhances the model's performance by incorporating domain-specific knowledge into the prediction process.

The HWAD layer is initialized with a JSON file containing size comparison rules and their weight values generated by OpenAI LLM using prompt engineering technique. The model when started training initially loads the weights along with the rules and extracts conditions and class labels from the JSON structure. A dictionary mapping label indices to class names and vice versa is also created to facilitate easy lookup during the forward pass.

During the forward pass, the layer processes bounding boxes, labels, and logits. It first computes the widths and lengths of the bounding boxes. Then, for each bounding box, the layer checks if it meets certain size comparison conditions relative to other bounding boxes and their classes detected based on the LLM produced JSON. This layer adjusts the corresponding logits based on the comparison rules and updated true weight values.

\begin{algorithm}[htb]
\caption{Hybrid Weight Adjustment(HWAD) Method Based on Size Comparison }\label{alg:update_truth_value}
\begin{algorithmic}[1]

\STATE \textbf{Input:} Rule \( R \) in JSON, Object Detection Data \( D \), Update Factor \( \alpha \)
\STATE \textbf{Output:} Updated Truth Value \( P(T|E) \)

\STATE Extract \( R \) from JSON
\STATE Initialize counters: \( C_{\text{obj}} \gets 0 \), \( C_{\text{sat}} \gets 0 \), \( C_{\text{not\_sat}} \gets 0 \)

\FOR{each \( I \in D \)}
    \STATE \( O_I \gets I[\text{"objects"}] \)
    \STATE \( S_I \gets I[\text{"sizes"}] \)
    \IF{\text{"individual\_object"} $\in$ \( O_I \)}

        \STATE \( C_{\text{obj}} \gets C_{\text{obj}} + 1 \)
        \STATE \( j \gets \text{index of "individual\_object" in } O_I \)
        \STATE \( \text{size}_{\text{obj}} \gets S_I[j] \)
        \IF{\( \text{size}_{\text{obj}} \) satisfies \( R \)}
            \STATE \( C_{\text{sat}} \gets C_{\text{sat}} + 1 \)
        \ELSE
            \STATE \( C_{\text{not\_sat}} \gets C_{\text{not\_sat}} + 1 \)
        \ENDIF
    \ENDIF
\ENDFOR

\STATE Calculate likelihoods:
\STATE \( n \gets C_{\text{sat}} + C_{\text{not\_sat}} \)
\STATE \( P(E|T) \gets \frac{C_{\text{sat}}}{n} \)
\STATE \( P(E|\neg T) \gets \frac{C_{\text{not\_sat}}}{n} \)

\STATE Calculate total evidence probability:
\STATE \( P(E) \gets P(E|T) \cdot P(T) + P(E|\neg T) \cdot P(\neg T) \)

\STATE Update posterior probability using Bayes' Theorem:
\STATE \( P(T|E) \gets \frac{P(E|T) \cdot P(T)}{P(E)} \)
\STATE \( P(T|E) \gets (1 - \alpha) \cdot \text{initial\_llm\_weight} + \alpha \cdot P(T|E) \)

\STATE Output the updated truth value:
\STATE \textbf{Return} \( P(T|E) \)

\STATE Update the rule \( R \) in JSON with \( P(T|E) \)

\STATE Reset counters:
\STATE \( C_{\text{obj}} \gets 0 \), \( C_{\text{sat}} \gets 0 \), \( C_{\text{not\_sat}} \gets 0 \)
\end{algorithmic}
\end{algorithm}

Additionally, the layer includes functions to parse the JSON file and create a size mapping that relates vehicles to their size comparison rules. The calculate\_posterior function computes the posterior probability for each rule, taking into account the number of times a rule is satisfied or not satisfied. This posterior probability is used to update the truth value or weights in the JSON file, which is then saved for future use.

Listing \ref{lst:jsonld-example} in the appendix, illustrates an example of the weights determined by the LLM for two classes, considering size comparison rules relative to other classes from a common-sense perspective. Rather than relying solely on the LLM-generated weights, an update factor ($\alpha$) is employed in HWAD technique. This factor combines $\alpha$ times the dataset-derived weight ratio with the remaining proportion from the initial LLM-generated weight. This approach allows adjusting the weights by incorporating both the dataset's actual conditions and the LLM's knowledge base, ensuring the weight updates reflect both actual conditions and informed predictions made by LLM.
\section{Experimental Result and Discussion}
\subsection{Dataset}
The experiments were conducted using two relatively small, publicly available datasets. The first dataset was the \textbf{Cars From Drone Dataset (CDD)}\cite{cars-from-drone-f5kny_dataset}, which comprised 463 aerial images containing five classes of land vehicles: bicycle, motorcycle, car, bus, and truck. The second dataset, a \textbf{Drone Vehicle Dataset (DVD)}\cite{dronevehicle_dataset}, was comparatively larger, consisting of 17,927 images of land-based vehicles across five classes: bus, car, freight car, truck, and van.

In addition to these, we developed a novel dataset named the \textbf{Multi-Environmental Vehicle Dataset (MEVD)}, which includes six classes. This dataset incorporates the five classes from the Cars From Drone Dataset and adds 177 images of boats, sourced from another dataset\cite{visualization-tools-for-aerial-ship-detection-dataset}. We selected only the images containing boats, and their associated annotations were converted to the MS-COCO format. 

\subsection{Computational Setup and Experimental Design}

Different well-established rules are incorporated into the model as custom layers in our PGCNN framework including redundancy elimination layers, scene based CAWAL Weight Adjustment Layer and HWAD Weight Adjustment Layer mentioned in the methodology section. In our experiments, we employed a Faster R-CNN model with a ResNet-50 backbone, pre-trained on ImageNet. Our initial experiment focused on making our notion a reality by incorporating three physical rules in three custom layers—two for redundancy removal and one for HWAD based on size comparison into the PGCNN framework to detect land vehicles and water vehicles using the CDD and DVD dataset. We then incorporated the model with a contextual custom layer 
designed to identify different scenes in an image such as roads for land vehicles and water areas for water vehicles. A scene segmentation model was trained with U-Net architecture\cite{ronneberger2015u} and used to determine the scenes from the image. The models were fine-tuned with the same hyperparameters presented in table \ref{tab:hyperparameters} in appendix section. 

\subsection{Results }

\begin{figure}[!htbp]
    \centering
    \includegraphics[ height = 5cm]{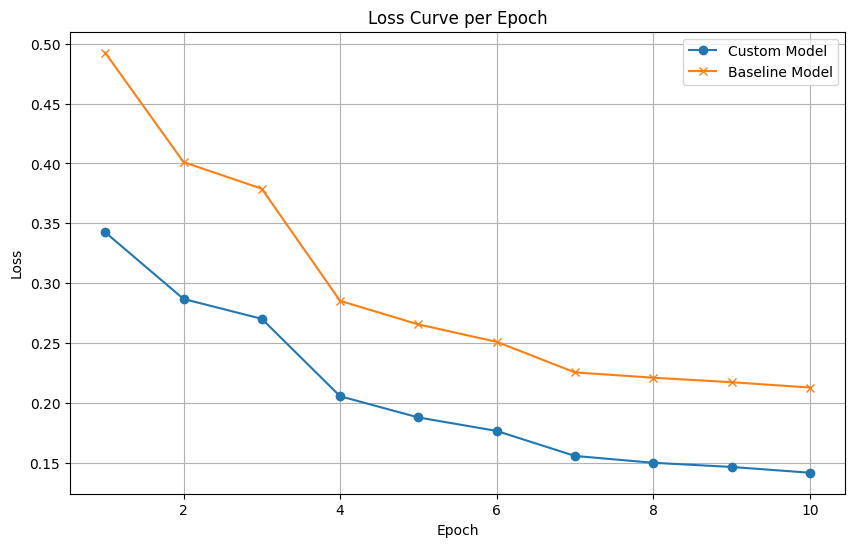}
    \caption{ Loss Curve of Baseline and PGCNN Framework on DVD dataset.}
    \label{fig:LossCurve}
\end{figure}

Table \ref{tab:metric} presents the mean average precision (mAP) and Intersection over Union (IoU) for the baseline model which represents a version of the model without any integrated physical rules, while the PGCNN model includes custom layers designed to incorporate these rules. The PGCNN model performs better than the baseline on the CDD dataset, with a mean Average Precision (mAP) of 0.4502 as opposed to 0.42. Furthermore, at thresholds of 0.5, 0.75, and 0.9, the PGCNN model exhibits improvements in Avg IoU, indicating more accurate bounding box predictions, especially at higher IoU thresholds. Conversely, on the DVD dataset, the PGCNN model shows a lower mAP (0.221) than the baseline (0.325), despite achieving better Avg IoU scores across all thresholds. This shows that even while the PGCNN can predict bounding boxes with more accuracy, additional work may need to be done to improve its overall detection performance on the DVD dataset. In MEVD almost all the scores are the same except for IoU at 0.9. Overall, the results indicate that the PGCNN model generally enhances detection performance, particularly in scenarios requiring higher precision. Figure \ref{fig:LossCurve} shows the downward trend of both the loss curve during training time for baseline and the PGCNN model on DVD dataset. Although the loss decreased over time in both models, PGCNN model exhibits a consistently lower loss compared to the baseline model across all epochs with significant reduction indicating improved convergence and potentially better generalization. The first loss obtained in base model started with 0.49 and ended with 0.21 while PGCNN started with 0.34 and the last epoch was 0.14. This demonstrates that the custom PGCNN model performs more effectively in minimizing loss during training compared to the baseline.
\begin{table*}[htbp]
\centering
\caption{Performance Comparison of Baseline and Custom PGCNN Models Across Test Datasets}
\label{tab:metric}
{%
\begin{tabular}{c|c|c|c|c|c|c}
\hline
\multicolumn{1}{l|}{}                                              & \multicolumn{2}{c|}{\textbf{CDD}}                                                                                                                   & \multicolumn{2}{c|}{\textbf{DVD}}                                                                                                                   & \multicolumn{2}{c}{\textbf{MEVD}}                                                                                                                 \\ \cline{2-7} 
\multicolumn{1}{l|}{\multirow{-2}{*}{\textbf{Evaluation Metrics}}} & \multicolumn{1}{c|}{\textbf{Baseline}} & \textbf{PGCNN} & \multicolumn{1}{c|}{\textbf{Baseline}} & \textbf{PGCNN} & \multicolumn{1}{c|}{\textbf{Baseline}} & \textbf{PGCNN} \\ 
\hline
mAP                                                                 & \multicolumn{1}{c|}{{ 0.420}}          & { \textbf{0.450}} & \multicolumn{1}{c|}{{ \textbf{0.325}}}          & { 0.221} & \multicolumn{1}{c|}{{ 0.218}} & { \textbf{0.218}} \\ 

{ Avg IoU @ 0.5}                                & \multicolumn{1}{c|}{{ 0.839}}          & { \textbf{0.851}}  & \multicolumn{1}{c|}{{ 0.804}}          & { \textbf{0.813}} & \multicolumn{1}{c|}{{ 0.758}} & { \textbf{0.758}} \\ 

{ Avg IoU @ 0.75}                               & \multicolumn{1}{c|}{{ 0.881}}          & { \textbf{0.903}}  & \multicolumn{1}{c|}{{ 0.856}}          & { \textbf{0.858}} & \multicolumn{1}{c|}{{ 0.869}} & { \textbf{0.869}} \\ 

{ Avg IoU @ 0.9}                                & \multicolumn{1}{c|}{{ 0.920}}          & { \textbf{0.926}}  & \multicolumn{1}{c|}{{ 0.926}}          & { \textbf{0.927}} & \multicolumn{1}{c|}{{ 0.926}}          & { \textbf{0.929}} \\ 
\hline
\end{tabular}%
}
\end{table*}

\subsubsection{Mitigation of False Positives(FP)}
Minimizing false positive occurrences is essential for improving the accuracy and trustworthiness of any model. When compared to the baseline model, the PGCNN model performs better in two key areas:

\textbf{Effects of IoU Metrics on Boundary Box Minimizing} For inference and to evaluate model performance, we selected images using a random image picker from each respective test dataset. We can see from Table \ref{tab:Bbox Reduction}, that the PGCNN model with custom layers outperforms the baseline model. As previously observed from the IoU results, the PGCNN model has improved in every threshold limit, indicating a decrease in overlapping and, consequently, a reduction in the number of boxes. Across the three datasets, the PGCNN framework achieved a substantial decrease in redundant bounding box detections, with the greatest reduction of 37.88\% observed on the CDD dataset despite being relatively smaller compared to the other two.
\begin{table}[h!]
\centering
\caption{Boundary Box Reduction Comparison of Baseline and PGCNN}
\label{tab:Bbox Reduction}
\begin{tabular}{c|cc|c}
\hline
                          & \multicolumn{2}{c|}{\textbf{Boundary Box detection}}                          &                                  \\      
                          \cline{2-3}

\multirow{-2}{*}{\textbf{Dataset}} & \multicolumn{1}{c|}{\textbf{Baseline}} & \textbf{PGCNN} & \multirow{-2}{*}{\textbf{Reduction(\%)}} \\ \hline
CDD                       & \multicolumn{1}{c|}{598}                                       & \textbf{451}                           & 37.88                            \\ 

%\hline
MEVD                      & \multicolumn{1}{c|}{909}                                       & \textbf{726}                           & 34.21                              \\ 

%\hline
DVD                       & \multicolumn{1}{c|}{ 6192}                                         & \textbf{5548}                                     & 10.56                                   \\ 

\hline
\end{tabular}%
\end{table}

\textbf{Mislabeled Boundary Box Reduction Impact:} If we look at table \ref{tab:FPcomparison} we can see determining false positives using the baseline and custom model on randomly picked 78 images from the test data presents a clear distinction of the reduction in mislabeled false positive detection using PGCNN. The base model identified 110 false positives for water vehicle classes, while the custom model reduced this number to 28. Similarly, for land vehicle classes, the base model detected 182 false positives, whereas the custom model reduced this to 111. This comparison highlights the effectiveness of PGCNN in significantly reducing the number of false positives for both water and land vehicles improving inference results. 

\begin{table}[H]
   \centering
   
    \caption{Identifying Number of False Positives Obtained from Images for Land and Water Vehicles in MEVD}

    \begin{tabular}{l|c|c|c}
        \hline
        \textbf{Class} & \textbf{Base Model} & \textbf{PGCNN} & \textbf{Reduction(\%)} \\
        \hline
        Water & 110 & 28 & 74.55\\
        \hline
        Land & 182 & 111 & 39.01\\
        \hline
    \end{tabular}
    \label{tab:FPcomparison}
\end{table}

\begin{figure}[h!]
    \centering
    \caption{Example Image of Mislabeled FP Reduction}
    \begin{subfigure}{0.47\textwidth}
        \centering
        \includegraphics[width=7cm,height=5cm]{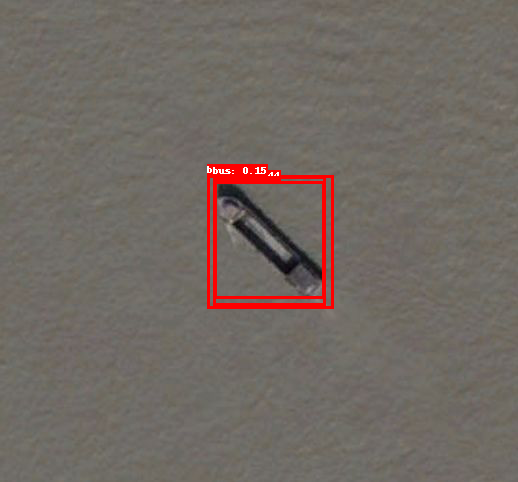}
        \caption{Baseline model predicts 2 bboxes with ‘bus’ labels for a ‘Boat’ object.}
        \label{fig:bbox_with_boat}
    \end{subfigure}
    \hfill
    \begin{subfigure}{0.47\textwidth}
        \centering
        \includegraphics[width=7cm,height=5cm]{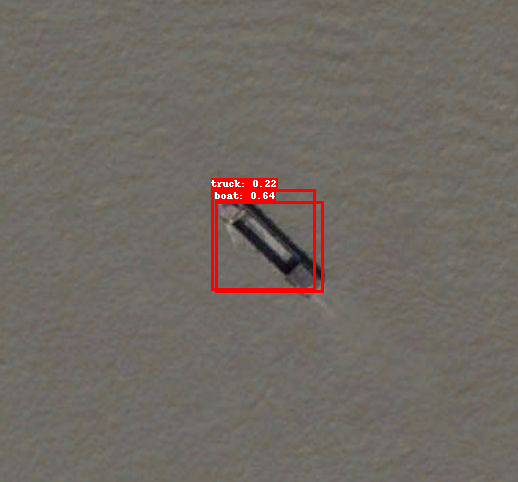}
        \caption{Custom Model detects ‘Boat’ class with a high prediction score and only 1 FP ‘truck’ with a low score.}
        \label{fig:truck_fp1}
    \end{subfigure}
    \label{fig:horizontal7}
\end{figure}

The PGCNN model significantly reduced false positive detections for water vehicle classes, achieving a \textbf{74.55\% }decrease. For instance, in images containing boats in the water area, the model was less likely to mistakenly identify land vehicles such as cars or trucks as false positives. Similarly, for land vehicle classes, the model demonstrated a \textbf{39.01\%} reduction in false positive occurrences. In this case, the model was less prone to incorrectly detecting water vehicles like boats in images featuring land vehicles such as cars or buses on roads. Overall \textbf{52.4\%} reduction in false positives is observed based on the MEVD dataset. This experiment was conducted on the MEVD dataset to fully understand the impact of all the physical rules implemented as custom layers in the model, including the CAWAL layer based on scenes. Figure \ref{fig:horizontal7} shows how PGCNN reduces the mislabeled detections based on scene context in a watermass.

\subsubsection{Optimizing Confidence Scores:}
The PGCNN model focused on optimizing the confidence scores during training, which resulted in improved confidence scores compared to the baseline model. High confidence scores for accurate detections can enhance the model's reliability and precision. To test the performance of the prediction or confidence score we randomly selected 78 images from the MEVD test set which had both land and water vehicles. 
% Detection's were labeled false positive, when the overlapping of the boundary box with ground truth was less than 60\% threshold value and another when the class label was detected  incorrectly. 
\begin{table}[ht]
\centering
\caption{Change in Confidence Scores Based on Physical Rules Using MEVD and PGCNN Framework}

\label{tab:ConfidenceScore}
{%
\begin{tabular}{|l|c|c|}
\hline
\textbf{Metric}          & \textbf{Score($\uparrow$)} & \textbf{Score($\downarrow$)} \\ \hline
\textbf{No. of Samples}   & 395              & 430              \\ 
\textbf{Percentage (\%)} & 66.05            & 71.91            \\ \hline
\end{tabular}%
}
\end{table}

A total of 430 false positive detections were identified by our PGCNN model, where confidence scores were reduced based on criteria such as less than 60\% overlap with the bounding box of different classes in the same area, the prohibition of two objects of the same class sharing the same bounding area, and physical attribute considerations such as context and size comparisons using the rules provided during training. Confidence score of detections that did not meet these conditions were reduced.
% such as cases where correctly labeled boxes had less than 60\% overlapping with ground truth and thus were detected with lower confidence values.
\begin{figure*}[htb]
     \centering
       
    \caption{Inference of Reduced Confidence Scores in FP}
     \begin{subfigure}{0.47\textwidth}
         \centering
         \includegraphics[width=7cm,height=5cm]{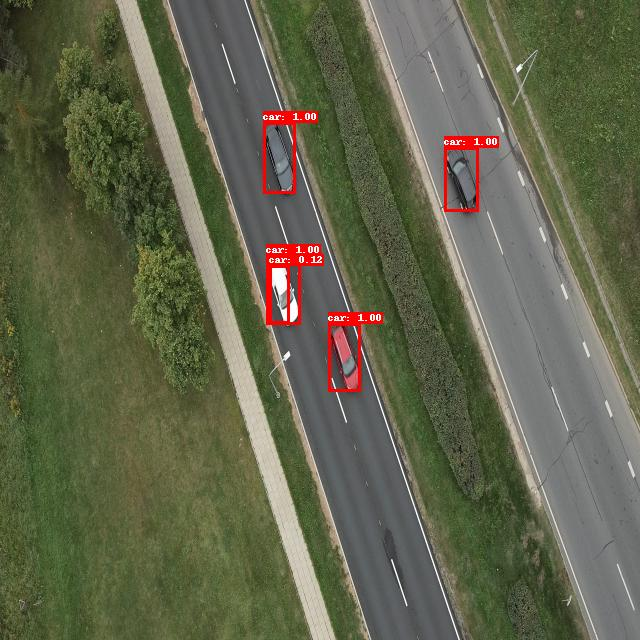}
        \caption{Baseline model predicts two cars: one with high confidence and another redundant but accurate with a confidence score of 0.12.}
         \label{fig:bboxw with boat }
     \end{subfigure}
     \hfill
     \begin{subfigure}{0.47\textwidth}
         \centering
         \includegraphics[width=7cm,height=5cm]{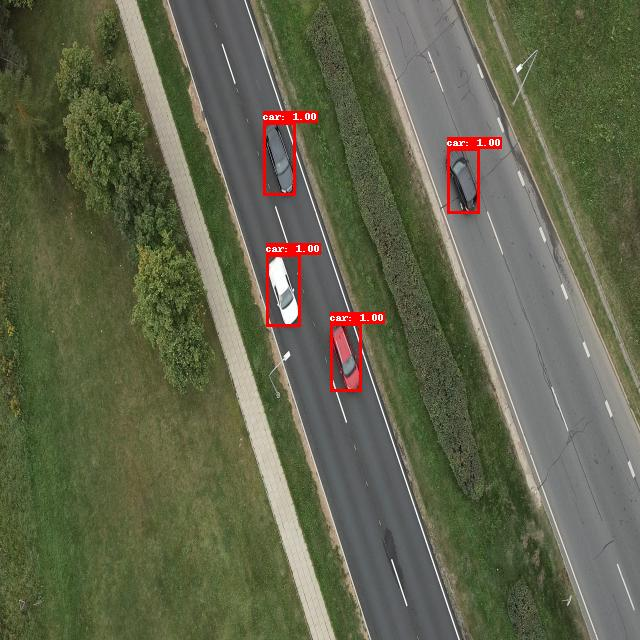}
         \caption{Custom model removes the redundant car Boundary box from the image.}
         \label{fig:tructfp1}
     \end{subfigure}

    \label{fig:horizontal8}
\end{figure*}
From figure \ref{fig:horizontal8} we see that, with a confidence level of 0.12, the baseline model predicts two cars one with high confidence and the other redundant but accurate. However, the custom PGCNN lowers the scores as a result the redundant car bounding box is removed from the image.
 
\begin{figure}[h!]
     \centering
       
    \caption{Demonstration of Updated Confidence Scores}
     \begin{subfigure}{0.47\textwidth}
         \centering
         \includegraphics[width=7cm,height=5cm]{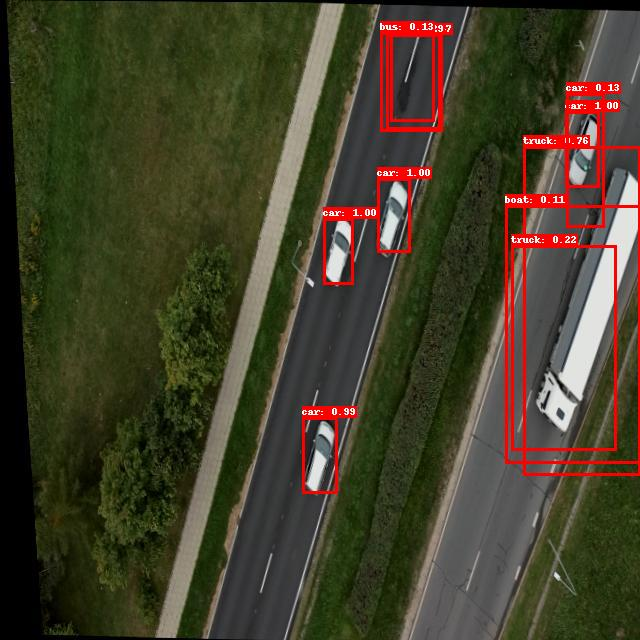}
         \caption{Baseline model predicts truck with highest 0.76 score \hspace{2pt} and more redundant boxes.}
         
         \label{fig:bboxw with boat }
     \end{subfigure}
     \hfill
     \begin{subfigure}{0.47\textwidth}
         \centering
         \includegraphics[width=7cm,height=5cm]{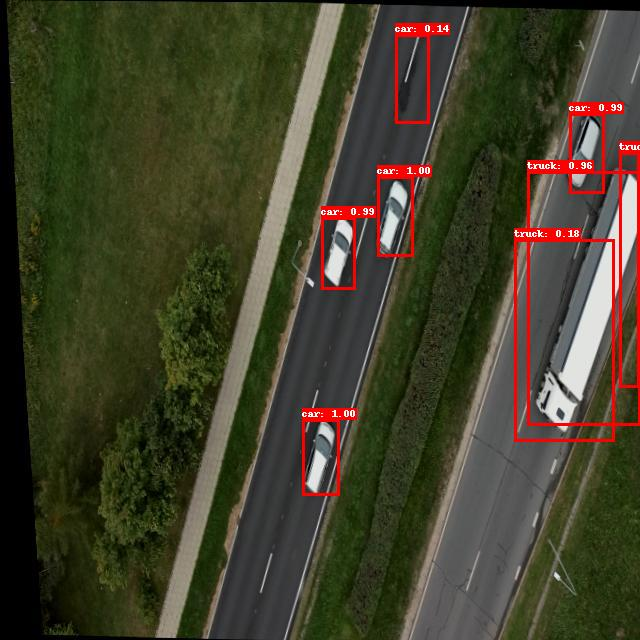}
         \caption{Custom model detects truck with 0.96 score \hspace{2pt}and the cars with either 1 or 0.99.}
         \label{fig:tructfp1}
     \end{subfigure}
    \label{fig:horizontal9}
\end{figure}

Moreover, 395 cases were detected where confidence scores was increased for accurate predictions based on the Physical rule conditions mentioned earlier which was implemented using the custom layers. Figure \ref{fig:horizontal9} shows the baseline model predicted the truck with the highest confidence score of 0.76 and accurate labeled data but resulting in more redundant boxes. PGCNN custom model detected the truck with a confidence score of 0.96 and identified the cars with scores of either 1 or 0.99 improving performance.

The PGCNN model demonstrated a \textbf{66.05}\% improvement in the confidence score of true positive detections relative to the baseline. Additionally, it reduced the confidence scores for redundant true positive detections significantly by \textbf{71.91}\% compared to the baseline. 

\section{Conclusion and Future Works}
The primary goal for our proposed PGCNN framework was to
upgrade the baseline CNN in way that it enhances reliability and accuracy. Our proposed PGCNN framework marks the beginning of a more reliable and trustworthy approach by blending traditional feature learning with real-world physical rules. Unlike conventional CNN architectures, PGCNN does not rely solely on learned features; instead, it verifies each prediction against predefined humanly-perceived conditions integrated into the framework, allowing the model to learn from and adapt to these conditions. This integration of rule-based modifications significantly reduces false positives, increases true positive accuracy, and minimizes redundant bounding boxes across all datasets used. This pioneering approach not only improves the model's precision and reliability but also represents the first instance of incorporating rule-based modifications into a CNN network, paving the way for more dependable applications across diverse domains such as speech, robotics, healthcare, and natural language processing. Specially with areas where more reliable models are ought to be implemented.

%Bibliography
\bibliographystyle{unsrt}  
%\bibliography{references} 

\begin{thebibliography}{10}

\bibitem{fu2023high}
Xiaotong Fu, Xiangyu Meng, Jing Zhou, and Ying Ji.
\newblock High-risk factor prediction in lung cancer using thin ct scans: An attention-enhanced graph convolutional network approach.
\newblock In {\em 2023 IEEE International Conference on Bioinformatics and Biomedicine (BIBM)}, pages 1905--1910. IEEE, 2023.

\bibitem{isufi2019generalizing}
Elvin Isufi, Fernando Gama, and Alejandro Ribeiro.
\newblock Generalizing graph convolutional neural networks with edge-variant recursions on graphs.
\newblock In {\em 2019 27th European Signal Processing Conference (EUSIPCO)}, pages 1--5. IEEE, 2019.

\bibitem{yaman2020self}
Burhaneddin Yaman, Seyed Amir~Hossein Hosseini, Steen Moeller, Jutta Ellermann, K{\^a}mil U{\u{g}}urbil, and Mehmet Ak{\c{c}}akaya.
\newblock Self-supervised learning of physics-guided reconstruction neural networks without fully sampled reference data.
\newblock {\em Magnetic resonance in medicine}, 84(6):3172--3191, 2020.

\bibitem{yaman2021ground}
Burhaneddin Yaman, Seyed Amir~Hossein Hosseini, Steen Moeller, Jutta Ellermann, K{\^a}mil U{\u{g}}urbil, and Mehmet Ak{\c{c}}akaya.
\newblock Ground-truth free multi-mask self-supervised physics-guided deep learning in highly accelerated mri.
\newblock In {\em 2021 IEEE 18th International Symposium on Biomedical Imaging (ISBI)}, pages 1850--1854. IEEE, 2021.

\bibitem{yaman2022self}
Burhaneddin Yaman.
\newblock {\em Self-Supervised Physics-Guided Deep Learning for Solving Inverse Problems in Imaging}.
\newblock PhD thesis, University of Minnesota, 2022.

\bibitem{huang2022physics}
Zhou Huang, Xinfeng Yin, and Yang Liu.
\newblock Physics-guided deep neural network for structural damage identification.
\newblock {\em Ocean Engineering}, 260:112073, 2022.

\bibitem{clifton2001sentence}
Charles Clifton~Jr and Susan~A Duffy.
\newblock Sentence and text comprehension: Roles of linguistic structure.
\newblock {\em Annual Review of Psychology}, 52(1):167--196, 2001.

\bibitem{biber2019text}
Douglas Biber.
\newblock Text-linguistic approaches to register variation.
\newblock {\em Register studies}, 1(1):42--75, 2019.

\bibitem{apple1979effects}
William Apple, Lynn~A Streeter, and Robert~M Krauss.
\newblock Effects of pitch and speech rate on personal attributions.
\newblock {\em Journal of personality and social psychology}, 37(5):715, 1979.

\bibitem{zhang2019pgnn}
Yongbing Zhang, Yangzhe Liu, Xiu Li, Shaowei Jiang, Krishna Dixit, Xinfeng Zhang, and Xiangyang Ji.
\newblock Pgnn: Physics-guided neural network for fourier ptychographic microscopy.
\newblock {\em arXiv preprint arXiv:1909.08869}, 2019.

\bibitem{banerjee2023physics}
Chayan Banerjee, Kien Nguyen, Clinton Fookes, and George Karniadakis.
\newblock Physics-informed computer vision: A review and perspectives.
\newblock {\em arXiv preprint arXiv:2305.18035}, 2023.

\bibitem{ling2024physics}
Hang~Jung Ling, Salom{\'e} Bru, Julia Puig, Florian Vix{\`e}ge, Simon Mendez, Franck Nicoud, Pierre-Yves Courand, Olivier Bernard, and Damien Garcia.
\newblock Physics-guided neural networks for intraventricular vector flow mapping.
\newblock {\em IEEE Transactions on Ultrasonics, Ferroelectrics, and Frequency Control}, 2024.

\bibitem{li2023physics}
Huihui Li, Linfeng Gou, Huacong Li, and Zhidan Liu.
\newblock Physics-guided neural network model for aeroengine control system sensor fault diagnosis under dynamic conditions.
\newblock {\em Aerospace}, 10(7):644, 2023.

\bibitem{fernandez2023physics}
Juan Fern{\'a}ndez, Juan Chiach{\'\i}o, Manuel Chiach{\'\i}o, Jos{\'e} Barros, and Matteo Corbetta.
\newblock Physics-guided bayesian neural networks by abc-ss: Application to reinforced concrete columns.
\newblock {\em Engineering Applications of Artificial Intelligence}, 119:105790, 2023.

\bibitem{zhang2023physics}
Weichang Zhang, Rui Zhang, Guoqing Wang, Wei Li, Xun Liu, Yang Yang, and Die Hu.
\newblock Physics guided remote sensing image synthesis network for ship detection.
\newblock {\em IEEE Transactions on Geoscience and Remote Sensing}, 61:1--14, 2023.

\bibitem{singh2018dock}
Krishna~Kumar Singh, Santosh Divvala, Ali Farhadi, and Yong~Jae Lee.
\newblock Dock: Detecting objects by transferring common-sense knowledge.
\newblock In {\em Proceedings of the European Conference on Computer Vision (ECCV)}, pages 492--508, 2018.

\bibitem{cars-from-drone-f5kny_dataset}
VilniusTech University.
\newblock Cars from drone dataset.
\newblock \url{ https://universe.roboflow.com/vilniustech-university/cars-from-drone-f5kny }, nov 2023.
\newblock visited on 2024-08-02.

\bibitem{dronevehicle_dataset}
Mranmay Shetty.
\newblock Dronevehicle dataset.
\newblock \url{ https://universe.roboflow.com/mranmay-shetty/dronevehicle }, may 2023.
\newblock visited on 2024-08-13.

\bibitem{visualization-tools-for-aerial-ship-detection-dataset}
Dataset Ninja.
\newblock Visualization tools for ship detection from aerial images dataset.
\newblock \url{ https://datasetninja.com/aerial-ship-detection }, aug 2024.
\newblock visited on 2024-08-02.

\bibitem{ronneberger2015u}
Olaf Ronneberger, Philipp Fischer, and Thomas Brox.
\newblock U-net: Convolutional networks for biomedical image segmentation.
\newblock In {\em Medical image computing and computer-assisted intervention--MICCAI 2015: 18th international conference, Munich, Germany, October 5-9, 2015, proceedings, part III 18}, pages 234--241. Springer, 2015.

\bibitem{qin2020u2}
Xuebin Qin, Zichen Zhang, Chenyang Huang, Masood Dehghan, Osmar~R Zaiane, and Martin Jagersand.
\newblock U2-net: Going deeper with nested u-structure for salient object detection.
\newblock {\em Pattern recognition}, 106:107404, 2020.

\bibitem{kirillov2023segment}
Alexander Kirillov, Eric Mintun, Nikhila Ravi, Hanzi Mao, Chloe Rolland, Laura Gustafson, Tete Xiao, Spencer Whitehead, Alexander~C Berg, Wan-Yen Lo, et~al.
\newblock Segment anything.
\newblock In {\em Proceedings of the IEEE/CVF International Conference on Computer Vision}, pages 4015--4026, 2023.

\end{thebibliography}

% \appendix

% \section{Appendix A: Additional Experimental Results}
\section{Appendix A: Shape-Based Object Detection Layer} 
In this study, we incorporated a shape-based object detection layer into our Physics-Guided Convolutional Neural Network (PGCNN) model to improve its performance for vehicle detection. We started by segmenting the Region of Interest (ROI) for each detected object, then upscaling the ROI regions with image enhancement algorithms to capture detailed features. We started by segmenting the Region of Interest (ROI) for each detected object, then upscaling the ROI regions with image enhancement algorithms to capture detailed features.To isolate the objects from their backgrounds, we applied the U-2Net\cite{qin2020u2} model, which made the backgrounds transparent. The transparent ROIs were then segmented using the SAM (Segment Anything Model)\cite{kirillov2023segment} approach to produce segments for each ROI object. Subsequently, a shape detection model, trained on a geometric shapes dataset comprising five classes (square, triangle, rectangle, parallelogram, and trapezoid), was employed to predict the shapes of these segments. This shape detection model was fine-tuned for 20 epochs using a pre-trained model with ReLU activation and a softmax output. During the training phase, we utilized a prompt generated by OpenAI LLM to predict the number and types of shapes that could be present in vehicle segments. These informations were stored in a JSON format which was used during training time to extract information and update the weights using HWAD technique. If the predicted shapes matched the expected number and types, the logit scores were updated to reflect increased confidence; otherwise, the target bounding box was removed from consideration. The following prompt was used to generate the shape predictions:

% \begin{quote}
% What shapes are among these classes: \texttt{'square', 'triangle', 'rectangle', 'parallelogram', 'trapezoid'} can be found when I see a bus, truck, car, motorcycle, and bicycle from a bird's-eye view? Also, provide their count. Imagine images are taken from satellite or drone.
% \end{quote}
\noindent\texttt{What shapes are among these classes: 'square', 'triangle', 'rectangle', 'parallelogram', 'trapezoid' can be found when I see a bus, truck, car, motorcycle, and bicycle from a bird's-eye view also provide their count imagine images are taken from satellite or drone.}

\begin{figure*}[htb]
     \centering
       
    \caption{Example of Predicted Labels of Shape Detection on Inference Image}
     \begin{subfigure}{0.47\textwidth}
         \centering
         \includegraphics[width=6.5cm,height=5cm]{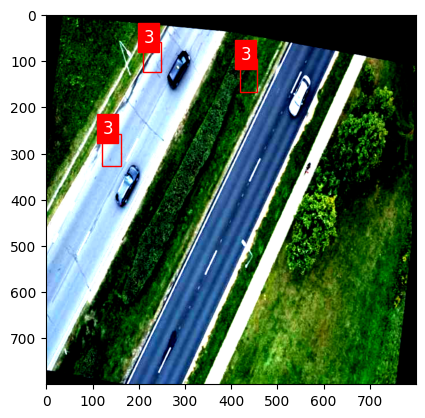}
         \label{fig:bboxShape1 }
     \end{subfigure}
     \hfill
     \begin{subfigure}{0.47\textwidth}
         \centering
         \includegraphics[width=6.5cm,height=5cm]{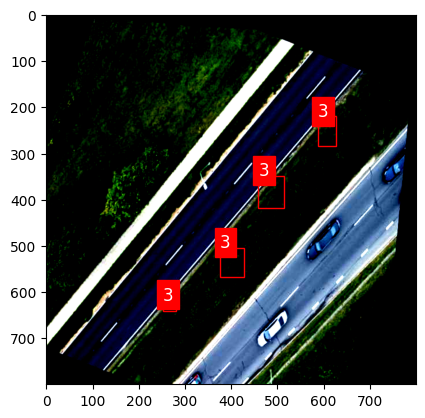}
         \label{fig:bboxShape2}
     \end{subfigure}

    \label{fig:bboxShape}
\end{figure*}

\begin{table}[h!]
\centering
\caption{LLM output for Shape Count of Different Vehicles from Bird's-Eye View}
\begin{tabular}{|c|c|c|c|c|}
\hline
\textbf{Vehicle} & \textbf{Rectangles} & \textbf{Squares} & \textbf{Trapezoids} & \textbf{Triangles} \\ \hline
Bus        & 1   & 0-1 & 2   & 0   \\ \hline
Truck      & 1-2 & 0-1 & 1   & 0   \\ \hline
Car        & 1   & 0-1 & 2   & 0   \\ \hline
Motorcycle & 1   & 0   & 0   & 1   \\ \hline
Bicycle    & 0-1 & 0   & 0   & 1-2 \\ \hline
\end{tabular}
\end{table}
% \begin{figure}[H]
%     \centering
%     \large
%     \includegraphics[width = 8cm,height = 4.5cm]{figures/ShapeMethodolog.drawio.png}
%     \caption{ Shape Segmentation Layer Overview}
%     \label{fig:shapeOverview}
% \end{figure}
\begin{figure}[H]
    \centering
    \large
    \includegraphics[width = 8cm,height = 5cm]{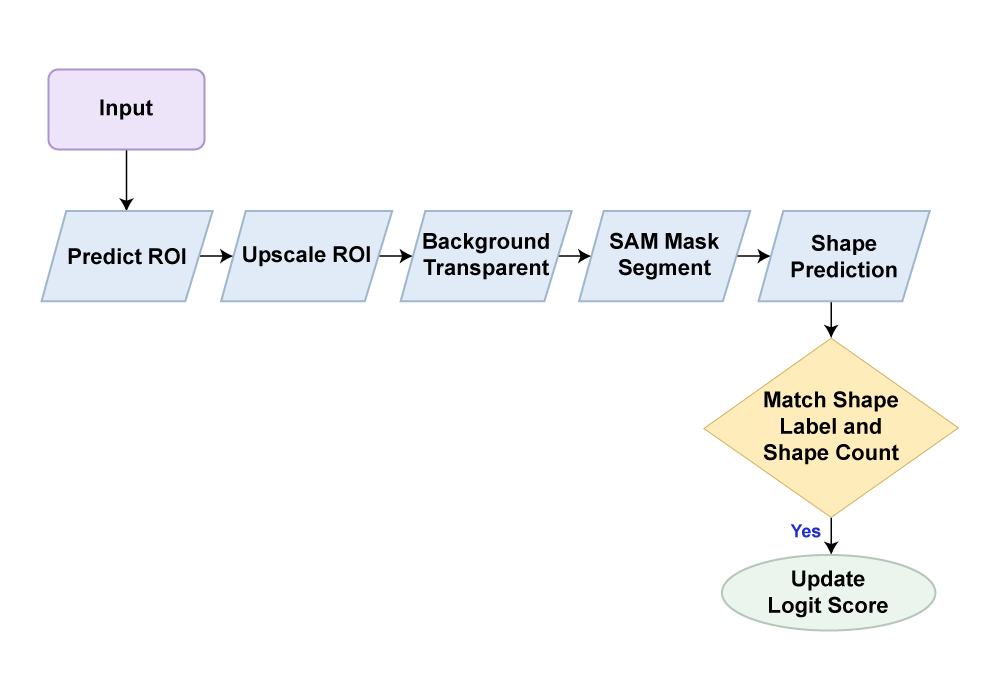}
    \caption{Shape Segmentation Layer Overview}
    \label{fig:shapeOverview}
\end{figure}
From figure \ref{fig:bboxShape} we can see that despite our efforts to detect objects with improve results adding this custom layer resulted in decreased accuracy for the RoI predictions. Ongoing invesigation is being done on how to improve this result and make the layer more efficient for detection. 

\section{Appendix B: Supplementary Data and Setup }
\label{appendix:a}

We used a device with linux (Ubuntu 22.04.2 LTS) operating system specifications with 256 GB of RAM and an AMD Ryzen Threadripper 3970x 32-core processor. Contains 4 Quadro RTX 5000 GPUs, each with 16 GB of memory. OpenAI was employed for common sense true weight value generation based on rules provided. Table \ref{tab:hyperparameters} represents the hyper parameters and others factors used in the experimental:

\begin{table}[H]
\centering
\caption{Training Hyperparameters}
\label{tab:hyperparameters}
\begin{tabular}{l|c}
\hline
\textbf{Hyperparameter} & \textbf{Value} \\ \hline
Model Backbone          & ResNet-50 (pretrained) \\ \hline
% Number of Classes       & 7 (including background) \\ \hline
Optimizer               & SGD \\ 
%\hline
Learning Rate           & 0.005 \\ 
%\hline
Momentum                & 0.9 \\ 
%\hline
Weight Decay            & 0.0005 \\ 
%\hline
Learning Rate Scheduler & StepLR \\  
%\hline
Gamma                   & 0.1 \\ 
%\hline
Number of Epochs        & 20 \\ 
%\hline
\hline
RF  (Redundant Factor)     & 60\% \\ 
%\hline
$\Delta t$  (CAWAL Threshold)     & 30\% \\ 
%\hline
$\alpha $ (HWAD Update Factor)     & 50\% \\ 
\hline
\end{tabular}
\end{table}
% \begin{figure}[H]
%     \centering
%     \large
%     \includegraphics[width = 6cm,height = 6cm]{figures/JsonSnippet2.png}
%     \caption{ Example Snippet of Weights Generated by LLM For 'Motorcycle' and 'Bicycle' Class.}
%     \label{fig:JSONSnippet}
% \end{figure}

The initial prompt used to generate the size relation weights for the Hybrid Weight Adjustment (HWAD) Layer:

\noindent\texttt{Generate a JSON structure representing a knowledge graph with nodes for the classes 'Car', 'Bus', 'Truck', 'Bicycle', 'Motorcycle', and 'Boat'. Each node should have edges that indicate its relationship to other nodes using 'isSmallerThan' and 'isBiggerThan' properties, with associated weights. Ensure that the relationships reflect the relative sizes of these vehicles.
The range of the weights should be between 0 to 1.
}

Later we adjusted the prompt and the JSON structure was built based on the weights generated. Listing \ref{lst:jsonld-example} illustrates a snippet of the weights determined by the LLM for two classes.
\begin{lstlisting}[language=XML, label={lst:jsonld-example}]
{     "@graph": [
      {
        "@id": "/c/en/Background",
        "@type": "node",
        "subClassOf": "Vehicle",
        "edges": []
      },
      {
        "@id": "/c/en/Car",
        "@type": "node",
        "subClassOf": "Vehicle",
        "edges": [{
            "isSmallerThan": [
              { "@id": "/c/en/Bus", "weight": 0.97 },
              { "@id": "/c/en/Truck", "weight": 0.93 }
            ],
            "isBiggerThan": [
              { "@id": "/c/en/Motorcycle", "weight": 0.94 },
              { "@id": "/c/en/Bicycle", "weight": 0.95 }
            ]
          }
        ]
      },
      {
        "@id": "/c/en/Bus",
        "@type": "node",
        "subClassOf": "Vehicle",
        "edges": [{
            "isBiggerThan": [
              { "@id": "/c/en/Truck", "weight": 0.50 },
              { "@id": "/c/en/Car", "weight": 0.97 },
              { "@id": "/c/en/Motorcycle", "weight": 1 },
              { "@id": "/c/en/Bicycle", "weight": 1 }
            ]
          }
        ]
      },
      {
        "@id": "/c/en/Boat",
        "@type": "node",
        "subClassOf": "Vehicle",
        "edges": [{
            "isBiggerThan": [
              { "@id": "/c/en/Bus", "weight": 0.80 },
              { "@id": "/c/en/Truck", "weight": 0.85 },
              { "@id": "/c/en/Car", "weight": 0.90 },
              { "@id": "/c/en/Motorcycle", "weight": 0.95 },
              { "@id": "/c/en/Bicycle", "weight": 0.95 }
            ] 
          }
        ]
      }
    ]
  }
  
\end{lstlisting}

\section{Appendix C: Dynamic Layer Integration Challenges} 

\subsection{CAWAL- Addressing Misclassification in Scene Segmentation}
\begin{figure}[H]
    \centering
    \large
    \includegraphics[width = 12cm,height = 4cm]{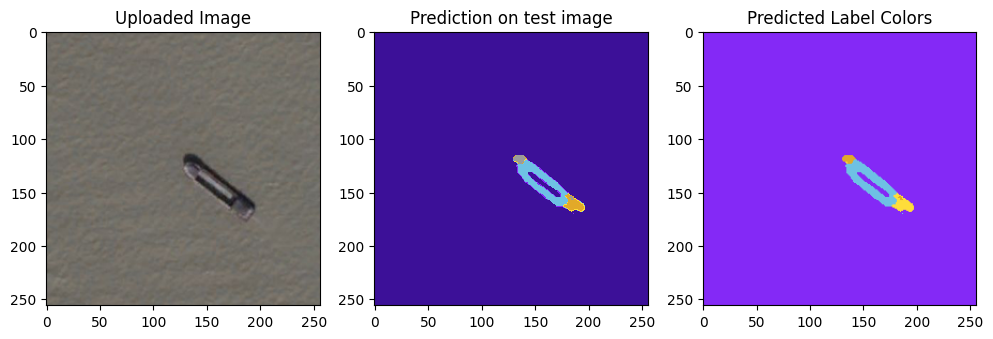}
    \caption{Example Image of Murky Water Detected as Road }
    \label{fig:murkywater}
\end{figure}

One of the significant issue we faced after training the scene segmentation model for the CAWAL layer was the frequent misinterpretation of murky water images as roadways. This issue was mostly caused by the color similarity between the water and road surfaces, which resulted in inaccurate labeling during segmentation as seen in figure \ref{fig:murkywater}. The differences between the training dataset used for scene label prediction and our three datasets impacted this, exposing the model's sensitivity to changes in the input data. We plan to explore alternative approaches in the future, such as fine-tuning the model with a more relevant dataset, to address these misclassification issues and increase overall segmentation accuracy.

\subsection{Impact of Class Imbalance on Model Performance in MEVD}

\begin{figure}[H]
    \centering
    \large
    \includegraphics[width = 10cm,height = 4cm]{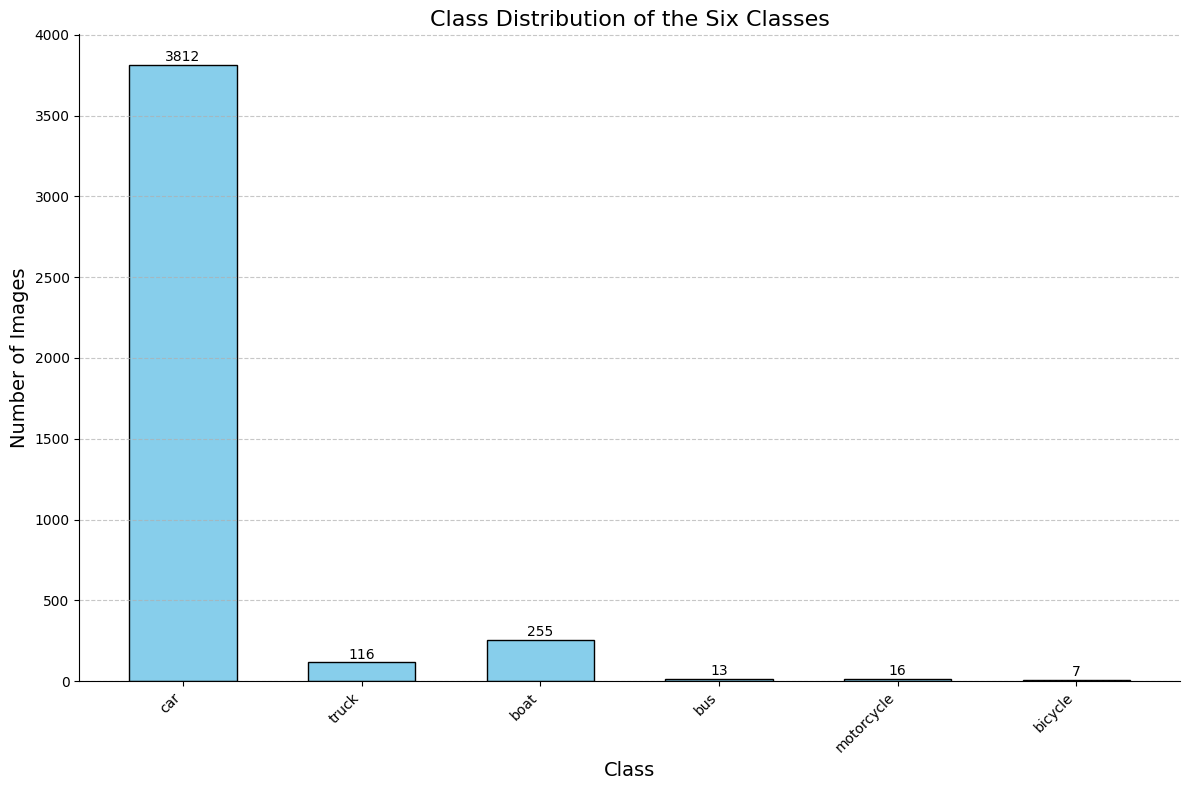}
    \caption{Imbalanced Class Distribution in MEVD Dataset}
    \label{fig:mevdLabel}
\end{figure}

The MEVD dataset, which combines the CDD dataset with boat images from another dataset, showed a large class imbalance, as depicted in the figure \ref{fig:mevdLabel}. The 'car' had tclass the most images(3812),  while other classes, such as 'bus' and 'boat', had much fewer. This imbalance caused PGCNN model to be biased towards car class. The model struggled to generalize successfully and thus impacted the overall performance in the MEVD dataset, not allowing the dynamic layers to be fully effective in adapting during training time. Although our PGCNN model outperformed the baseline, this limitation hindered its ability to accurately predict minority classes and fully utilize the dynamic layers, resulting in suboptimal outcomes.

\end{document}